\documentclass[runningheads]{llncs}

\usepackage{url}
\usepackage{graphicx}
\usepackage{bm}
\usepackage{amsmath}
\usepackage{amssymb}
\usepackage{amsfonts}
\usepackage{booktabs}
\usepackage{subfigure}
\usepackage{multirow}
\usepackage{pifont}

\newcommand*{\affaddr}[1]{#1} 
\newcommand*{\affmark}[1][*]{\textsuperscript{#1}}

\title{DCA: Diversified Co-Attention towards Informative Live Video Commenting}
\author{Zhihan Zhang\affmark[1], Zhiyi Yin\affmark[1], Shuhuai Ren\affmark[2], Xinhang Li\affmark[3] and Shicheng Li\affmark[1]
}
\institute{
\affaddr{\affmark[1]School of Electronic Engineering and Computer Science, Peking University, China}\\
\affaddr{\affmark[2]School of Software Engineering, Huazhong University of Science and Technology, China}\\ 
\affaddr{\affmark[3]College of Software, Beijing University of Aeronautics and Astronautics, China}
\email{\{zhangzhihan, yinzhiyi\}@pku.edu.cn, 
renshuhuai007@gmail.com}\\
\email{hestiaskylee@gmail.com, lisc99@pku.edu.cn}
}
\authorrunning{Z. Zhang, Z. Yin, S. Ren, X. Li and S. Li} 

\begin{document}
\maketitle
\begin{abstract}
We focus on the task of Automatic Live Video Commenting (ALVC), which aims to generate real-time video comments with both video frames and other viewers' comments as inputs. A major challenge in this task is how to properly leverage the rich and diverse information carried by video and text. In this paper, we aim to collect diversified information from video and text for informative comment generation. To achieve this, we propose a Diversified Co-Attention (DCA) model for this task. Our model builds bidirectional interactions between video frames and surrounding comments from multiple perspectives via metric learning,  to collect a \textit{diversified} and \textit{informative} context for comment generation. We also propose an effective parameter orthogonalization technique to avoid excessive overlap of information learned from different perspectives.  Results show that our approach outperforms existing methods in the ALVC task, achieving new state-of-the-art results.
\end{abstract}

\section{Introduction}
\begin{figure}[tb]
\setlength{\belowcaptionskip}{-0.5cm}
    \centering
    \includegraphics[width=0.6\linewidth]{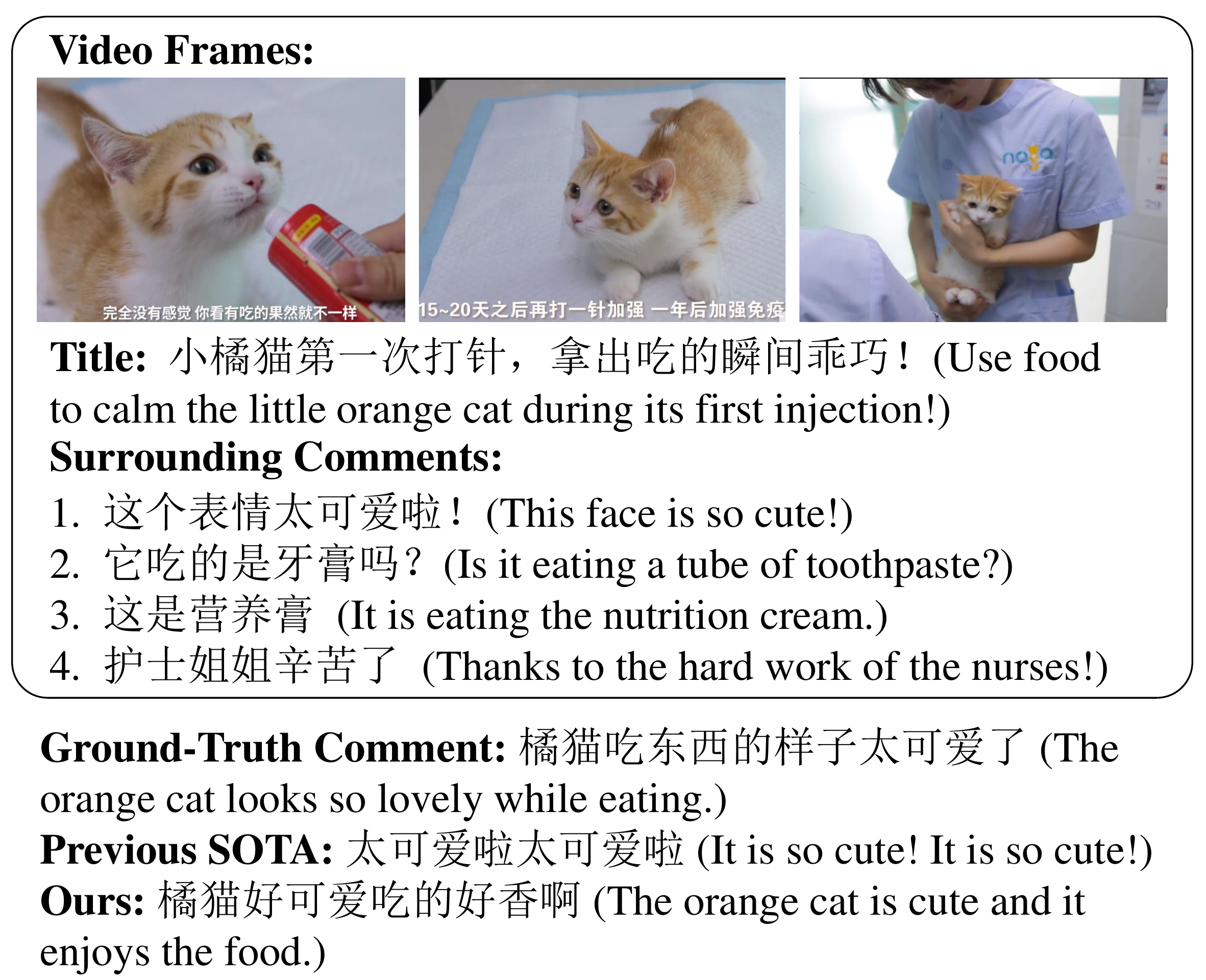}
    \caption{An example of the ALVC task. The inputs are video frames and surrounding comments. The ground-truth comment (written by human) is the desired output. Compared to previous SOTA, our model generates comment with richer information from both video frames (\textit{the orange cat}) and surrounding comments (\textit{it is eating}).}
    \label{fig:case}
\end{figure}

Live video commenting, also known as Danmaku commenting, is an emerging interaction mode among online video websites \cite{chen2017watching}. This technique allows viewers to write real-time comments while watching videos, in order to express opinions about the video or to interact with other viewers. Based on the features above, the \textbf{A}utomatic \textbf{L}ive \textbf{V}ideo \textbf{C}ommenting (ALVC) task aims to generate live comments for videos, while considering both the video and the surrounding comments made by other viewers. Figure \ref{fig:case} presents an example for this task. Automatically generating real-time comments brings more fun into video watching and reduces the difficulty of understanding video contents for human viewers. Besides, it also engages people's attention and increases the popularity of the video.

Despite its usefulness described above, the ALVC task has not been widely explored. Ma et al.\cite{ma2019livebot} is the first to propose this task, which is the only endeavor so far. They employ separate attention on the video and surrounding comments to obtain their representations. Such approach does not integrate visual and textual information and may lead to a limited information diversity. In fact, the surrounding comments are written based on the video, while they also highlight important features of the video frames. Thus, we aim to collect \textit{diversified} information from video and text by building interactions between these two modalities.

As an effective method in multi-modal scenarios, co-attention has been applied in multiple tasks~\cite{lu2016coatten,biatten2016,qanet,li2019beyond}. Based on previous works, we propose a novel \textbf{D}iversified \textbf{C}o-\textbf{A}ttention (DCA) model to better capture the complex dependency between video frames and surrounding comments. By learning different distance metrics to characterize the dependency between two information sources, the proposed DCA can build bidirectional interactions between video frames and surrounding comments from multiple perspectives, so as to produce \textit{diversified} co-dependent representations. Going a step further, we propose a simple yet effective parameter orthogonalization technique to avoid excessive overlap (\emph{information redundancy}) of information extracted from different perspectives. Experiment results suggest that our DCA model outperforms the previous approaches as well as the traditional co-attention, reaching state-of-the-art results in the ALVC task. Further analysis supports the effectiveness of the proposed components, as well as the information diversity in the DCA model.

\vspace{-0.05in}
\section{Diversified Co-Attention Model}
\begin{figure}[tb]
\setlength{\belowcaptionskip}{-0.3cm}
\centering
\includegraphics[width=0.95\linewidth]{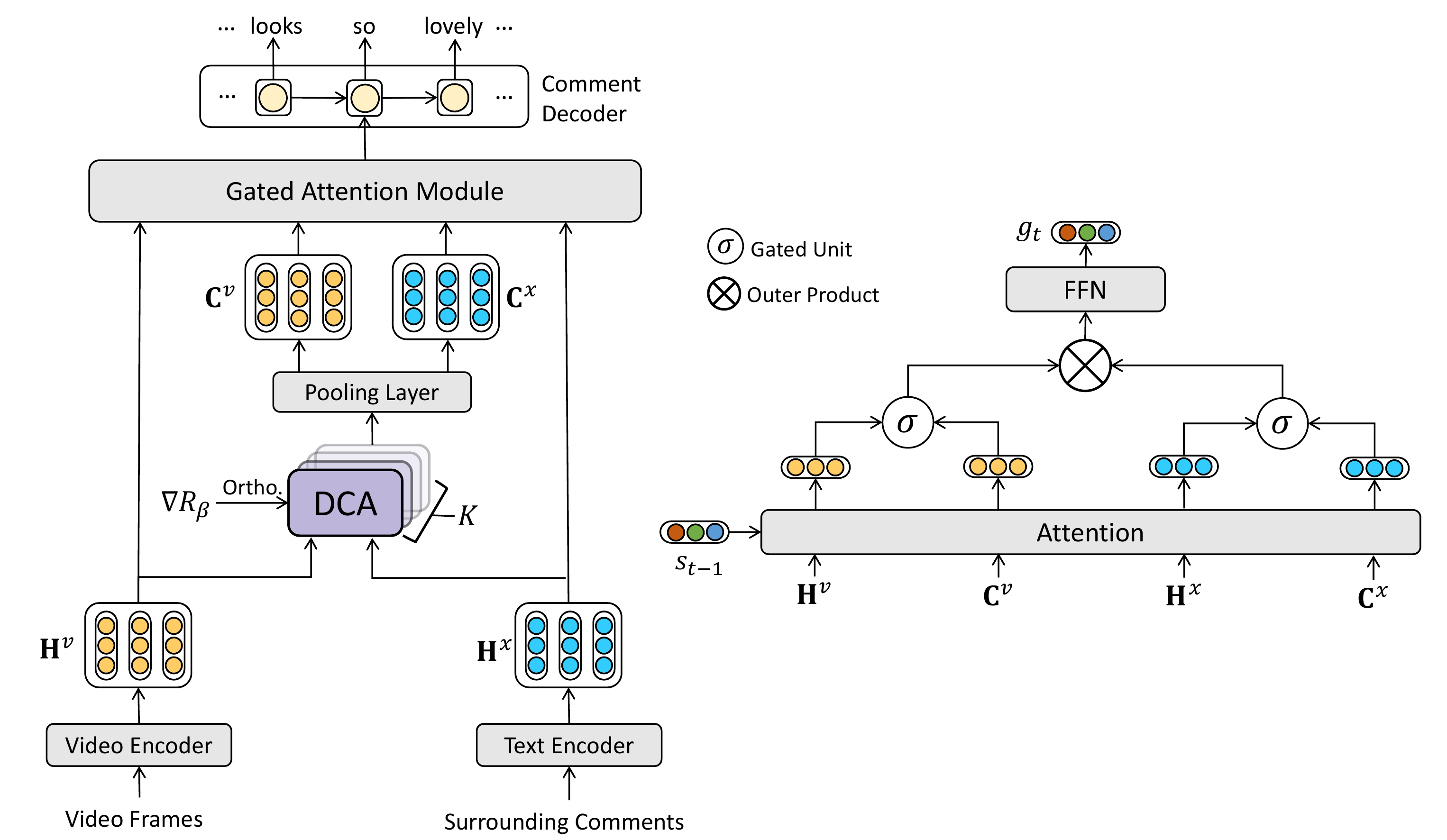}
\caption{An overview of the proposed DCA model (left) and the details of gated attention module (right).}
\label{fig:model}
\end{figure}

\vspace{-0.05in}
Given video frames $\bm{v}=(v_1,\cdots,v_n)$ and surrounding comments $\bm{x}=(x_1,\cdots,x_m)$\footnote{We concatenate all surrounding comments into a single sequence $\bm{x}$.}, the ALVC task aims at generating a reasonable and fluent comment $\bm{y}$. Figure \ref{fig:model} presents the sketch of our DCA model.

\subsection{Video Encoder \& Text Encoder}
\vspace{-0.1cm}

The video encoder and text encoder aim to obtain representations of video frames and surrounding comments, respectively. The encoders are implemented as GRU networks. The hidden representations of each video frame $v_i$ and each word $x_i$ is computed as:
\vspace{-0.05in}

\begin{equation}
\label{eq:video_encoder}
    \bm{h}_i^v = {\rm GRU}\big(\bm{h}_{i-1}^v, f(v_i)\big),\quad
    \bm{h}_i^x = {\rm GRU}\big(\bm{h}_{i-1}^x, e(x_i)\big)
\end{equation}
where $f(\cdot)$ refers to a convolutional neural network (CNN) used to transform raw images into dense vectors, and $e(x_i)$ is the word embedding of $x_i$. The visual and textual representation matrices are denoted as $\mathbf{H}^v=\{\bm{h}_1^v,\cdots,\bm{h}_{n}^v\}\in\mathbb{R}^{n \times d}$ and $\mathbf{H}^x=\{\bm{h}_1^x,\cdots,\bm{h}_m^x\}\in\mathbb{R}^{m \times d}$, respectively. Here we assume that $\bm{h}_i^v$ and $\bm{h}_i^x$ share the same dimension. Otherwise, a linear transformation can be introduced to ensure that their dimensions are the same.

\vspace{-0.3cm}
\subsection{Diversified Co-Attention}
\vspace{-0.1cm}
\label{sec:dca}

To effectively capture the dependency between video frames and surrounding comments, we propose a \textbf{D}iversified \textbf{C}o-\textbf{A}ttention (DCA) mechanism, which builds bidirectional interactions between two sources of information from different perspectives via metric learning~\cite{metriclearning}. We first elaborate on our DCA from \emph{single perspective} and then extend it to \emph{multiple perspectives}.

\vspace{-0.05in}
\subsubsection*{Single Perspective} 
The single-perspective DCA is adapted from original co-attention~\cite{lu2016coatten}, but introduces techniques of metric learning. We first connect video representations $\mathbf{H}^v$ and text representations $\mathbf{H}^x$ by computing similarity between them. Conventionally, the similarity score between two vectors can be calculated as their inner product. However, the model is expected to learn a \textit{\textbf{task-specific distance metric}} in the joint space of video and text. Therefore, the similarity matrix $\mathbf{S}\in\mathbb{R}^{n \times m}$ between $\mathbf{H}^v$ and $\mathbf{H}^x$ is calculated as:

\vspace{-0.05in}
\begin{equation}
\label{eq:similarity}
\mathbf{S} = \mathbf{H}^v\mathbf{W}(\mathbf{H}^x)^{\rm T}
\end{equation}

where $\mathbf{W}\in\mathbb{R}^{d\times d}$ is a learnable parameter matrix. Here we constraint $\mathbf{W}$ as a \emph{positive semidefinite matrix} to ensure that Eq.~(\ref{eq:similarity}) satisfies the basic definition~\cite{distance} of the distance metric. Since $\mathbf{W}$ is continuously updated during model training, the \emph{positive semidefinite} constraint is difficult to keep satisfied. To remedy this, we adopt an alternative: $\mathbf{H}^v$ and $\mathbf{H}^x$ are first applied with the same linear transformation $\mathbf{L}\in\mathbb{R}^{d\times d}$, and then the inner product of transformed matrices is computed as their similarity score:

\vspace{-0.05in}
\begin{equation}
\label{eq:linear}
    \mathbf{S} = (\mathbf{H}^v\mathbf{L})(\mathbf{H}^x\mathbf{L})^{\rm T} = \mathbf{H}^v\mathbf{L}\mathbf{L}^{\rm T}(\mathbf{H}^x)^{\rm T}
\end{equation}

where $\mathbf{L}\mathbf{L}^{\rm T}$ can be regarded as an approximation of $\mathbf{W}$ in Eq.~(\ref{eq:similarity}). Since $\mathbf{L}\mathbf{L}^{\rm T}$ is \emph{symmetric positive definite}, it is naturally a \emph{positive semidefinite matrix}. Each element $\mathbf{S}_{ij}$ denotes the similarity score between $v_i$ and $x_j$. $\mathbf{S}$ is normalized row-wise to produce vision-to-text attention weights $\mathbf{A}^x$, and column-wise to produce text-to-vision attention weights $\mathbf{A}^v$. The final representations are computed as the product of attention weights and original features:

\vspace{-0.3cm}
\begin{align}
\label{eq:attention}
     \mathbf{A}^x = {\rm softmax}(\mathbf{S}),&\quad
     \mathbf{A}^v = {\rm softmax}(\mathbf{S^{\rm T}})\\
    \mathbf{C}^x = \mathbf{A}^x\mathbf{H}^x,&\quad
    \mathbf{C}^v = \mathbf{A}^v\mathbf{H}^v
\end{align}

where $\mathbf{C}^v\in\mathbb{R}^{m\times d}$ and $\mathbf{C}^x\in\mathbb{R}^{n\times d}$ denote the co-dependent representations of vision and text. Since $\mathbf{H}^v$ and $\mathbf{H}^x$ guide each other's attention, these two sources of information can mutually boost for better representations.

\subsubsection*{Multiple Perspectives} 

As distance metrics between vectors can be defined in various forms, learning a single distance metric $\mathbf{L}$ does not suffice to comprehensively measure the similarity between two kinds of representations. On the contrary, we hope to provide an informative context for the comment decoder from \textit{diversified} perspectives. To address this contradiction, we introduce a multi-perspective setting in the DCA. 

We ask the DCA to learn \textit{\textbf{multiple distance metrics}} to capture the dependencies between video and text from different perspectives. To achieve this, DCA learns $K$ different parameter matrices $\{\mathbf{L}_1,\cdots,\mathbf{L}_K\}$ in Eq.(\ref{eq:linear}), where $K$ is a hyper-parameter denoting the number of perspectives. Intuitively, each $\mathbf{L}_i$ represents a learnable distance metric. Given two sets of representations $\mathbf{H}^x$ and $\mathbf{H}^v$, each $\mathbf{L}_i$ yields a similarity matrix $\mathbf{S}_i$ as well as co-dependent representations $\mathbf{C}^x_i$ and $\mathbf{C}^v_i$ from its unique perspective. DCA is then able to build
bi-directional interactions between two information sources from multiple perspectives. Finally, a mean-pooling layer is used to integrate the representations from different perspectives:
\vspace{-0.05in}
\begin{equation}
\label{eq:mean_pooling}
     \mathbf{C}^x = \frac{1}{K}\sum_{k=1}^K\mathbf{C}^x_k,\quad
     \mathbf{C}^v = \frac{1}{K}\sum_{k=1}^K\mathbf{C}^v_k
     \vspace{-0.05in}
\end{equation}

\subsection{Parameter Orthogonalization}
\label{sec:ortho}

One potential problem of the above multi-perspective setting is \textit{information redundancy}, meaning that the information extracted from different perspectives may overlap excessively. Specifically, the parameter matrices $\{\mathbf{L}_k\}_{k=1}^K$ may tend to be highly similar after many rounds of training. According to~\cite{se}, to alleviate this problem, $\{\mathbf{L}_k\}_{k=1}^K$ should be as orthogonal as possible. We first try to add a regularization term  $R_\beta$ into the loss function as an orthonormality constraint~\cite{Parseval}:
\begin{equation}
\setlength{\abovedisplayskip}{3pt}
\setlength{\belowdisplayskip}{3pt}
R_\beta = \frac{\beta}{4}\sum\limits_{i=1}^K\sum\limits_{j=1}^K\left({\rm tr}\left(\mathbf{L}_i\mathbf{L}_j^{\rm T}\right)-\mathbb{I}(i=j)\right)^2
\end{equation}
where ${\rm tr}(\cdot)$ is the trace of the matrix and $\beta$ is a hyper-parameter. However, we empirically find that the simple introduction of regularization term may cause the collapse of model training. Thus, we propose an approximate alternative: after back propagation updates all parameters at each learning step, we adopt a post-processing method equivalent to the aforementioned orthonormality constraint by updating $\{\mathbf{L}_k\}_{k=1}^K$ with the gradient of regularization term $R_\beta$:

\vspace{-0.5cm}
\begin{align}
\setlength{\abovedisplayskip}{3pt}
\setlength{\belowdisplayskip}{3pt}
\nabla_{\mathbf{L}_i}R_\beta =& \beta\left(\sum_{k=1}^K{\rm tr}\left(\mathbf{L}_i\mathbf{L}_k^{\rm T}\right)\mathbf{L}_k-\mathbf{L}_i\right)\\
\label{eq:post}
\mathbf{L}_i \xleftarrow[]{update} & (1+\beta)\mathbf{L}_i-\beta\sum_{k=1}^K{\rm tr}(\mathbf{L}_i\mathbf{L}_k^{\rm T})\mathbf{L}_k
\end{align}

The orthonormality constraint ensures that $\{\mathbf{L}_k\}_{k=1}^K$ are nearly orthogonal, suggesting that the information carried by these matrices rarely overlaps~\cite{se}. By reducing \textit{information redundancy} in the multi-perspective setting, the orthogonalization technique ensures the \textit{diversity} of information collected by DCA.

\subsection{Gated Attention Module}

In order to integrate the co-dependent representations from the DCA and original representations from the encoders, a Gated Attention Module (GAM) is designed following the DCA. Given the hidden state $\bm{s}_{t-1}$ of the decoder at timestep $t-1$, we first apply attention mechanism on the co-dependent and original representations respectively, using $\bm{s}_{t-1}$ as query:
\begin{equation}
\setlength{\abovedisplayskip}{5pt}
\setlength{\belowdisplayskip}{5pt}
\label{eq:gam_attn}
    \bm{\widehat{c}}^x = \mathcal{A}(\bm{s}_{t-1}, \mathbf{C}^x),\quad
    \bm{\widehat{h}}^x = \mathcal{A}(\bm{s}_{t-1}, \mathbf{H}^x)
\end{equation}
where $\mathcal{A}$ is the attention mechanism \cite{attention2014}. Then, $\bm{\widehat{c}}^x$ and $\bm{\widehat{h}}^x$ are passed through a gated unit to generate comprehensive textual representations:

\begin{equation}
\setlength{\abovedisplayskip}{5pt}
    \bm{w}^x = \sigma(\mathbf{U}_c^x\bm{\widehat{c}}^x + \mathbf{U}_h^x\bm{\widehat{h}}^x + b_x),\quad
    \label{eq:gate}
    \bm{r}^x = \bm{w^x}\odot\bm{\widehat{c}}^x + (\bm{1}-\bm{w^x})\odot\bm{\widehat{h}}^x
\end{equation}

where $\mathbf{U}_c^x$, $\mathbf{U}_h^x$ and $b_x$ are learnable parameters, $\sigma$ denotes the sigmoid function and $\odot$ denotes element-wise multiplication. $\bm{r}^x$ is the balanced textual representation of $\bm{\widehat{c}}^x$ and $\bm{\widehat{h}}^x$. Symmetrically, we obtain the balanced visual representation $\bm{r}^v$ through Eq.(\ref{eq:gam_attn})$\sim$Eq.(\ref{eq:gate}) based on $\mathbf{C}^v$ and $\mathbf{H}^v$.

In the ALVC task, the contribution of video information and textual information towards the desired comment may not be equivalent. Therefore, we calculate the final context vector $\bm{g}_t\in\mathbb{R}^{d}$ as:
\begin{equation}
\setlength{\abovedisplayskip}{5pt}
\setlength{\belowdisplayskip}{5pt}
    \bm{g}_t = \mathcal{FFN}\big(\bm{r}^x\otimes(\bm{\alpha}\odot\bm{r}^v)\big)
\end{equation}
where $\bm{\alpha}$ is a learnable vector. $\otimes$ denotes the outer product and $\mathcal{FFN}$ denotes a feed-forward neural network. The outer product is a more informative way to represent the relationship between vectors than the inner product, which we use to collect an informative context for generation.

\vspace{-0.3cm}
\subsection{Decoder}
\label{sec:comment_decoder}

Given the context vector $\bm{g}_t$ obtained by the GAM, the decoder aims to generate a comment $\bm{y}=(y_1,\cdots,y_l)$ via another GRU network. The hidden state $s_t$ at timestep $t$ is computed as:
\begin{equation}
\setlength{\abovedisplayskip}{3pt}
\setlength{\belowdisplayskip}{3pt}
\bm{s}_{t} = {\rm GRU}\big(\bm{s}_{t-1},\left[e(y_{t-1});\bm{g}_t\right]\big)
\end{equation}
where $y_{t-1}$ is the word generated at time-step $t-1$, and semicolon denotes vector concatenation. The decoder then samples a word $y_{t}$ from the output probability distribution:
\begin{equation}
\setlength{\abovedisplayskip}{3pt}
\setlength{\belowdisplayskip}{3pt}
y_{t} \sim {\rm softmax}(\mathbf{O}\bm{s}_{t})
\end{equation}
where $\mathbf{O}$ denotes an output linear layer. The model is trained by maximizing the log-likelihood of the ground-truth comment.

In order to test the universality of the proposed components, we also implement our model based on Transformer~\cite{transformer}. Specifically, the text encoder, video encoder and comment decoder are implemented as Transformer blocks. Since this extension is not the focus of this paper, we will not explain it in more detail. Readers can refer to~\cite{transformer} for detailed descriptions of the Transformer architecture.

\vspace{-0.05in}
\section{Experiments}
\label{sec:experiment}

\subsection{Data and Settings}

We conduct experiments on the Live Comment Dataset\footnote{https://github.com/lancopku/livebot} \cite{ma2019livebot}. The dataset is collected from the popular Chinese video streaming website Bilibili\footnote{https://www.bilibili.com}. It contains 895,929 instances in total, which belong to 2,361 videos. In experiments, we adopt 34-layer Resnet \cite{resnet} pretrained on ImageNet to process the raw video frames in Eq.(\ref{eq:video_encoder}). We set the number of perspectives to $K=3$ in Eq.~(\ref{eq:mean_pooling}) and $\beta$ in Eq.~(\ref{eq:post}) is set to 0.01.  We adopt the Adam \cite{Adam} optimization method with initial learning rate 3e-4, and train the model for 50 epochs with dropout rate 0.1.

\subsection{Baselines}

The baseline models in our experiments include the previous approaches in the ALVC task as well as the traditional co-attention model. For each listed Seq2Seq-based models, we implement another Transformer-based version by replacing the encoder and decoder to Transformer blocks.

\begin{itemize}
\vspace{-0.03in}
\item\textbf{S2S-Video}~\cite{show_and_tell} uses a CNN to encode the video frames and a RNN decoder to generate the comment. This model only takes the video frames as input.
\item\textbf{S2S-Text}~\cite{attention2014} is the traditional Seq2Seq model with attention mechanism. This model only takes the surrounding comments as input.
\item\textbf{S2S-Concat}~\cite{seq2seq-ic} adopts two encoders to encode the video frames and the surrounding comments, respectively. Outputs from two encoders are concatenated and fed into the decoder. 
\item\textbf{S2S-SepAttn}~\cite{ma2019livebot} employs separate attention on video and text representations. The attention contexts are concatenated and fed into the decoder.
\item\textbf{S2S-CoAttn} is a variant of our model, which replaces the DCA module using traditional co-attention \cite{lu2016coatten}.
\end{itemize}
Accordingly, the Transformer versions are named as \textbf{Trans-Video}, \textbf{Trans-Text}, \textbf{Trans-Concat}, \textbf{Trans-SepAttn} and \textbf{Trans-CoAttn}.

\vspace{-0.3cm}
\subsection{Evaluation Metrics}
\label{sec:metrics}

\vspace{-0.1cm}
\paragraph{Automatic evaluation}~{
Due to the \textit{diversity} of video commenting, we cannot collect all possible comments for reference-based comparison like BLEU. As a complement, rank-based metrics are applied in evaluating diversified generation tasks such as dialogue systems~\cite{das2017visual,duconv,kdconv}. Given a set of candidate comments, the model is asked to sort the candidates in descending order of likelihood scores. Since the model generates the sentence with the highest likelihood score, it is reasonable to discriminate a good model based on its ability to rank the ground-truth comment on the top. Following previous work \cite{ma2019livebot}, the 100 candidate comments are collected as follows: }
\label{para: rank-based}
\begin{itemize}
\item[$\diamond$]\textbf{Ground-truth:} 
The human-written comment in the original video.

\item[$\diamond$]\textbf{Plausible:} 
30 most similar comments to the video title in the training set. Plausibility is computed as the cosine similarity between the comment and the video title based on TF-IDF values. 

\item[$\diamond$]\textbf{Popular:} 
20 most frequently appeared comments in the training set, most of which are meaningless short sentences like ``Hahaha'' or ``Great''.

\item[$\diamond$]\textbf{Random:} 
Comments that are randomly picked from the training set to make the candidate set up to 100 sentences.
\end{itemize}

 We report evaluation results on the following metrics: \textbf{Recall@$\bm{k}$} (the percentage that the ground-truth appears in the top $k$ of the ranked candidates), \textbf{MR} (the mean rank of the ground-truth), and \textbf{MRR} (the mean reciprocal rank of the ground-truth).

\vspace{-0.3cm}
\paragraph{Human evaluation}
In human evaluation, we randomly pick 200 instances from the test set. We ask five human annotators to score the generated comments from different models on a scale of 1 to 5 (higher is better). The annotators are required to evaluate these comments from the following aspects: \textbf{Fluency} (whether the sentence is grammatically correct), 
\textbf{Relevance} (whether the comment is relevant to the video and surrounding comments), \textbf{Informativeness} (whether the comment carries rich and meaningful information) and
\textbf{Overall} (the annotator's general recommendation).

\begin{table*}[tb]
\caption{Results of automatic evaluation. \textbf{R@$\bm{k}$} is short for \textbf{Recall@$\bm{k}$}. Lower \textbf{MR} score means better performance, while other metrics are the opposite.}
\centering
\footnotesize
\setlength{\tabcolsep}{4pt}
\resizebox{1.0\columnwidth}!{
\begin{tabular}{l|c c c c c|l|c c c c c}
\toprule
\textbf{Seq2Seq}& \textbf{R@1} & \textbf{R@5} & \textbf{R@10} & \textbf{MRR} & \textbf{MR} & \textbf{Transformer} & \textbf{R@1} & \textbf{R@5} & \textbf{R@10} & \textbf{MRR} & \textbf{MR}\\
\midrule

S2S-Video & 4.7 & 19.9 & 36.5 & 14.5 & 21.6 & Trans-Video & 5.3 & 20.7 & 38.2 & 15.1 & 20.9 \\
S2S-Text & 9.1 & 28.1 & 44.3 & 20.1 & 19.8 & Trans-Text & 10.5 & 30.2 & 46.1 & 21.8 & 18.5\\
S2S-Concat & 12.9 & 33.8 & 50.3 & 24.5 & 17.1 & Trans-Concat & 14.2 & 36.8 & 51.5 & 25.7 & 17.2\\
S2S-SepAttn & 17.3 & 38.0 & 56.1 & 27.1 & 16.1 & Trans-SepAttn & 18.0 & 38.1 & 55.8 & 27.5 & 16.0\\
S2S-CoAttn & 21.9 & 42.4 & 56.6 & 32.6 & 15.5 & Trans-CoAttn & 23.1 & 42.8 & 56.8 & 33.4 & 15.6\\
\midrule
\textbf{DCA (S2S)} & \textbf{25.8} & \textbf{44.2} & \textbf{58.4} & \textbf{35.3} & \textbf{15.1} & \textbf{DCA (Trans)} & \textbf{27.2} & \textbf{47.6} & \textbf{62.0} & \textbf{37.7} & \textbf{13.9} \\
 \bottomrule
\end{tabular}
}
\label{tab:auto}
\end{table*}


\begin{table}[tb]
\caption{Results of human evaluation. We average the scores given by 5 annotators. Scores in bold indicate significant improvement ($\geqslant$ 0.5).}
\setlength{\belowcaptionskip}{-0.5cm}
\centering
\footnotesize
\setlength{\tabcolsep}{7.0pt}
\resizebox{0.75\columnwidth}!{
\begin{tabular}{l|c c c c}
\toprule
\textbf{Models} & \textbf{Fluency} & \textbf{Relevance} & \textbf{Informativeness} & \textbf{Overall} \\ 
 \midrule
S2S-Concat & 2.7 & 2.4 & 2.6 & 2.6 \\
S2S-SepAttn & 3.1 & 2.8 & 2.5 & 3.1 \\
S2S-CoAttn & 3.5 & 3.2 & 2.7 & 3.3 \\
\midrule
\textbf{DCA (S2S)} & 3.7 & 3.5 & \textbf{3.4} & 3.6\\
\midrule
Trans-Concat & 3.0 & 2.5 & 2.5 & 2.7 \\
Trans-SepAttn & 3.2 & 2.7 & 2.8 & 3.3 \\
Trans-CoAttn & 3.6 & 3.3 & 3.3 & 3.5 \\
\midrule
\textbf{DCA (Trans)} & 3.7 & 3.6 & \textbf{3.8} & 3.7\\
 \bottomrule
\end{tabular}
}

\label{tab:human}
\end{table}


\vspace{-0.3cm}
\subsection{Experiment Results}

According to the results of \textbf{automatic evaluation} (Table~\ref{tab:auto}), our DCA model assigns higher ranks to ground-truth comments. These results prove that DCA has stronger ability in discriminating highly relevant comments from irrelevant ones. Since the generation process is also retrieving the best sentence among all possible word sequences, it can be inferred that DCA performs better at generating high-quality sentences.

Additionally, our DCA model receives more favor from human judges in \textbf{human evaluation} (Table~\ref{tab:human}). This proves that DCA generates comments that are more consistent with human writing habits. We also discover that the margin between DCA and baselines in \textit{Informativeness} is larger than the other perspectives. Assisted by the proposed components to obtain diversified information from video and text, sentences generated by DCA are more informative than the other models.

The experiments show consistent results in Seq2Seq models and Transformer models. Hence, the proposed DCA modules are believed to have good \textbf{universality}, which can adapt to different model architectures.

\begin{table*}[t]
\caption{Experiment results of the ablation study.  \textbf{Ortho.} represents parameter orthogonalization. ``\textbf{-DCA}" means using traditional co-attention to replace DCA.}
\centering
\footnotesize
\setlength{\belowcaptionskip}{-0.8cm}
\resizebox{0.8\textwidth}!{
\begin{tabular}{l|c c c c c|c c c c c}
\toprule
 &
 \multicolumn{5}{c|}{\textbf{Seq2Seq Architecture}} & \multicolumn{5}{c}{\textbf{Transformer Architecture}}\\ 
\midrule
\textbf{Models}& \textbf{R@1} & \textbf{R@5} & \textbf{R@10} & \textbf{MRR} & \textbf{MR} & \textbf{R@1} & \textbf{R@5} & \textbf{R@10} & \textbf{MRR} & \textbf{MR}\\
\midrule
\textbf{Full Model} & \textbf{25.8} & \textbf{44.2} & \textbf{58.4} & \textbf{35.3} & \textbf{15.1} & \textbf{27.2} & \textbf{47.6} & \textbf{62.0} & \textbf{37.7} & \textbf{13.9} \\

\quad\textbf{-GAM}& 24.1 & 43.8 & 57.5 & 35.0 & 15.4 & 26.2 & 47.5 & 60.4 & 37.3 & 15.1 \\

\quad\textbf{-Ortho.}&  22.7 & 43.2 & 57.2 & 33.4 & 15.8 & 24.7 & 45.8 & 59.5 & 35.6 & 14.9 \\

\quad\textbf{-DCA} &21.9 & 42.4 & 56.6 & 32.6 & 15.5 & 23.1 & 42.8 & 56.8 & 33.4 & 15.6 \\

 \bottomrule
\end{tabular}}
\label{tab:incre}
\end{table*} 

\subsection{Ablation Study}
\label{sec:incre}

In order to better understand the efficacy of the proposed methods, we further conduct an ablation study on different settings of our model, with results presented in Table \ref{tab:incre}. 

As the results suggest, there is a significant drop in the model's performance while replacing the DCA module with traditional co-attention. Compared to traditional co-attention, DCA has advantages in its multi-perspective setting, \textit{i.e.}, learning multiple distance metrics in the joint space of video and text. DCA builds interactions between two information sources from multiple perspectives, hence extracting richer information than traditional co-attention.

Besides, results show that the parameter orthogonalization technique and the GAM module are also critical to our model's performance. By alleviating the \textit{information redundancy} issue in DCA's multi-perspective setting, the orthogonalization technique ensures the \textit{diversity} of information collected by DCA. GAM uses gated units to integrate information from co-dependent and original representations, as well as to balance the importance of video and text. Such approach helps GAM collect an informative context for comment generation.

\subsection{Visualization of DCA}
\label{sec:visual_dca}

\begin{figure}[!tb]
\setlength{\belowcaptionskip}{-0.3cm}
\centering
\subfigure[without orthogonalization]{\includegraphics[width=0.46\linewidth]{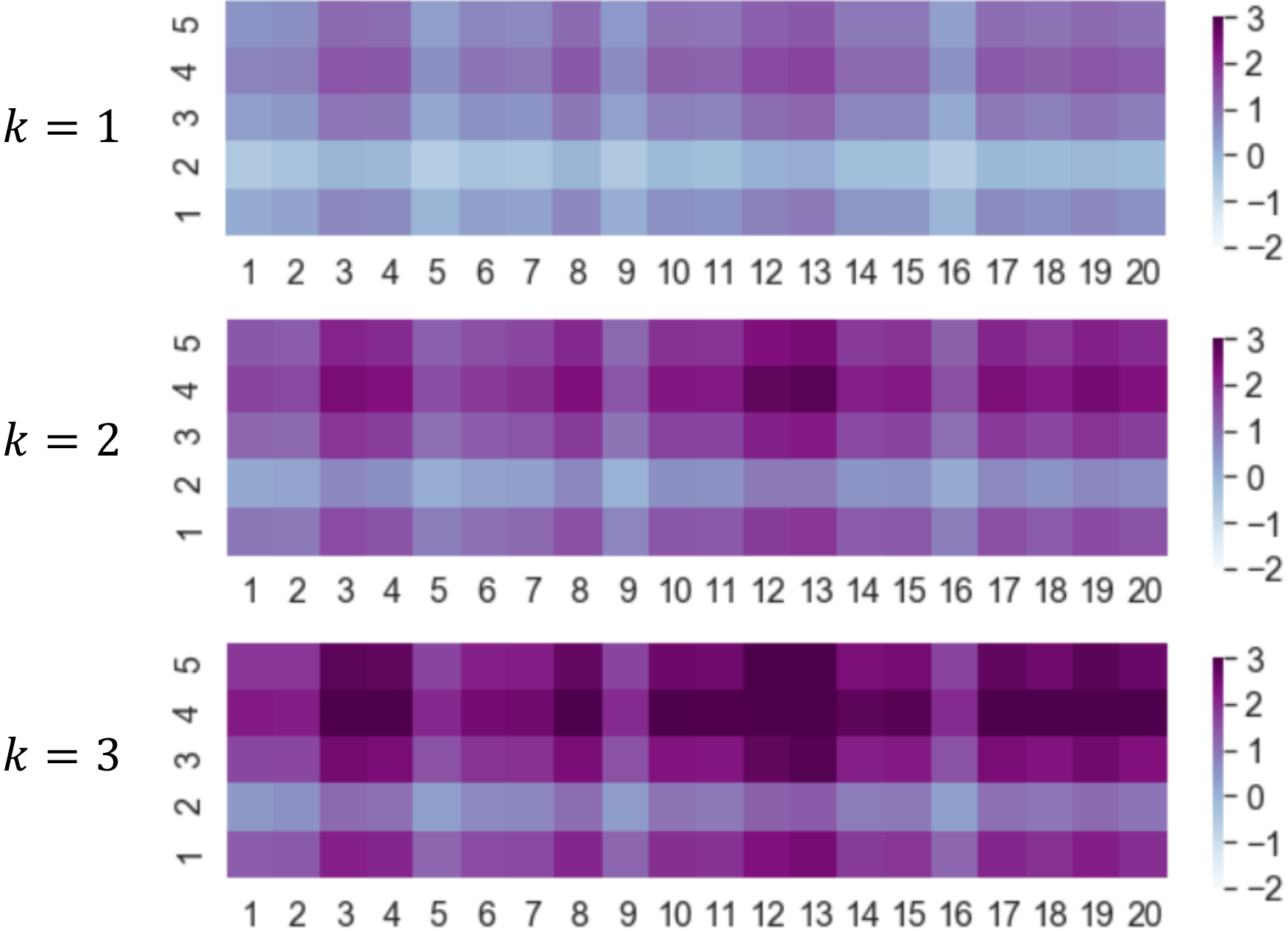}\label{fig-a}}
\subfigure[with orthogonalization ]{\includegraphics[width=0.46\linewidth]{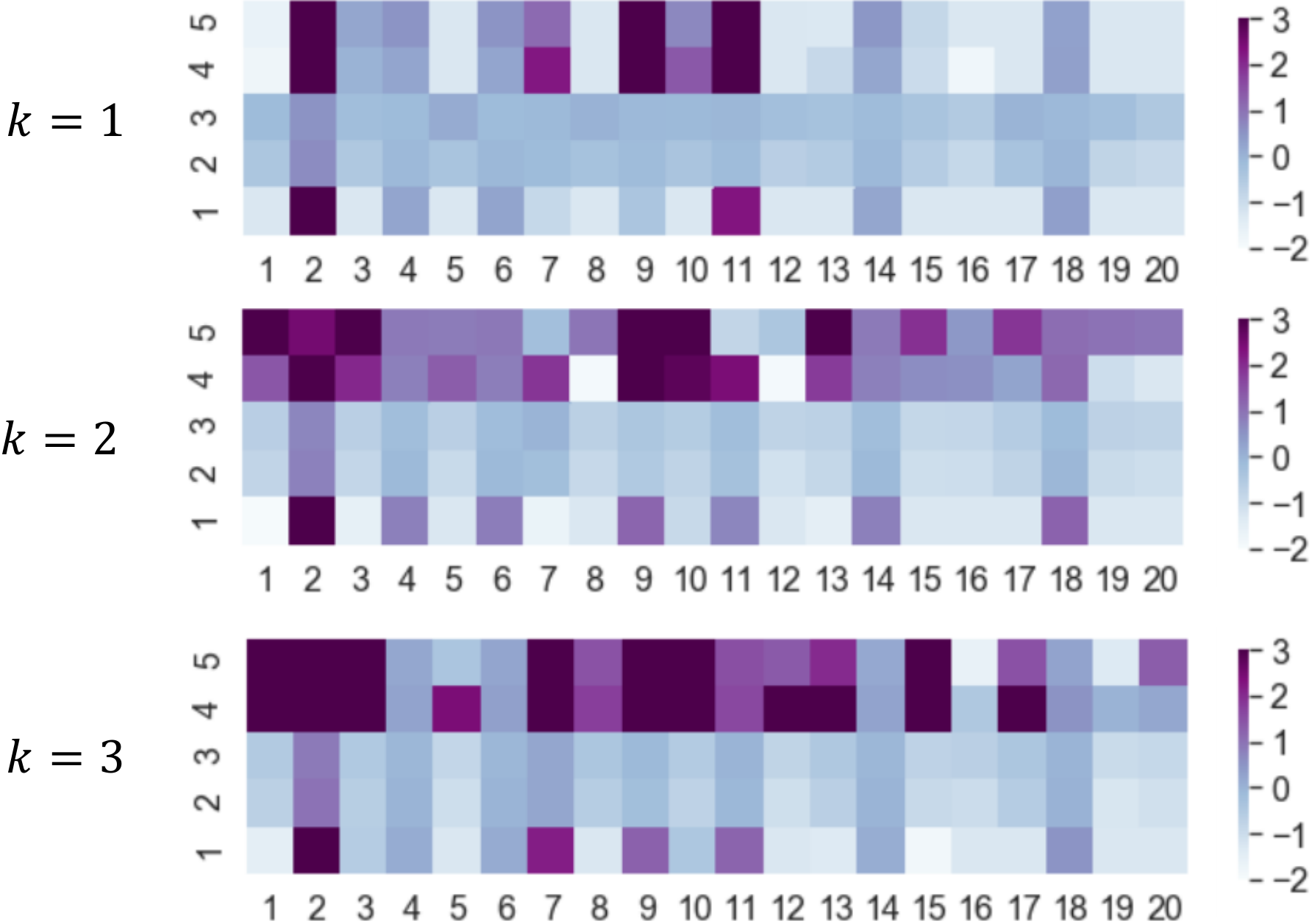}\label{fig-b}}
\caption{Visualization of similarity matrices in DCA with or without parameter orthogonalization. Here we set $K=3$. Horizontal axis: 20 words in a surrounding comment. Vertical axis: 5 surrounding video frames. Deeper color denotes higher relevance.}
\label{fig-postprocess}
\end{figure}

To illustrate the contribution of parameter orthogonalization to the information diversity of our model, we visualize the similarity matrices $\{\mathbf{S}_k\}_{k=1}^K$ in DCA. In the vanilla DCA (shown in Figure \ref{fig-postprocess}\subref{fig-a}), each $\mathbf{S}_i$ is generated by a distance metric $\mathbf{L}_i$ through Eq.(\ref{eq:linear}). However, the similarity matrices are highly similar to each other. This shows that the information extracted from $K$ perspectives suffers from the \textit{information redundancy} problem, which is consistent with our hypothesis in Section~\ref{sec:ortho}. After introducing the parameter orthogonalization (shown in Figure \ref{fig-postprocess}\subref{fig-b}), apparent differences can be seen among these similarity matrices. This further explains the performance decline after removing the orthogonalization technique in Table \ref{tab:incre}. The parameter orthogonalization ensures the discrepancy between distance metrics $\{\mathbf{L}_k\}_{k=1}^K$, helps DCA generate \textit{diversified} representations, thus alleviates information redundancy and improves information diversity.

\section{Related Work}
\vspace{-0.05in}
\paragraph{Automatic Article Commenting}
One similar task to our work is automatic article commenting. Qin et al.\cite{automatic_article_commenting} is the first to introduce this task and constructs a Chinese news dataset. Ma et al.\cite{ma2018comment} proposes a retrieval-based commenting framework on unpaired data via unsupervised learning. Yang et al.\cite{yang2019cross} leverages visual information for comment generation on graphic news. Zeng et al.\cite{zeng2019automatic} uses a gated memory module to generate personalized comment on social media. Li et al.\cite{li2019coherent} models the news article as a topic interaction graph and proposes a graph-to-sequence model. Compared to article commenting, the ALVC task aims to model the interactions between text and video, and video is a more dynamic and complex source of information. The co-dependent relationship between a video and its comments makes this task a larger challenge for AI models.

\vspace{-0.05in}
\paragraph{Video Captioning}
Another similar task to ALVC is video captioning. Venugopalan at al.\cite{seq2seq-ic} applies a unified deep neural network with CNN and LSTM layers. Shen et al.\cite{shen2017} proposes a sequence generation model with weakly supervised information for dense video captioning. Xiong et al.\cite{xiong2018move} produces descriptive paragraphs for videos via a recurrent network by assembling temporally localized descriptions. Li et al.\cite{li2019residual} uses a residual attention-based LSTM to reduce information loss in generation. Xu et al.\cite{xu2019joint} jointly performs event detection and video description via a hierarchical network. Compared to video description, the ALVC task requires not only a full understanding of video frames, but also interaction with other human viewers. This requires effective modeling of the intrinsic dependency between visual and textual information.

\vspace{-0.05in}
\paragraph{Co-Attention}
Our model is also inspired by the previous work of co-attention. Lu et al.\cite{lu2016coatten} introduces a hierarchical co-attention model in visual QA. Nguyen et al.\cite{nguyen2018improved} proposes a dense co-attention network with a fully symmetric architecture. Tay et al.\cite{tay2018multi} applies a co-attentive multi-pointer network to model user-item relationships. Hsu et al.\cite{hsu2018co} adds co-attention module into CNNs to perform unsupervised object co-segmentation. Yu et al.\cite{yu2019deep} applies a deep modular co-attention network in combination of self-attention and guided-attention. Li et al.\cite{li2019beyond} uses positional self-attention and co-attention to replace RNNs in video question answering. Compared to previous co-attention methods, DCA considers the issue of obtaining co-dependent representations as distance metric learning. Equipped with the parameter orthogonalization technique, DCA is able to obtain rich information from multiple perspectives.

\vspace{-0.05in}
\section{Conclusion}
\vspace{-0.05in}

This work presents a diversified co-attention model for automatic live video commenting to capture the complex dependency between video frames and surrounding comments. By introducing bidirectional interactions between the video and text from multiple perspectives (different distance metrics), two information sources can mutually boost for better representations. Besides, we propose an effective parameter orthogonalization technique to avoid excessive overlap of information extracted from different perspectives. Experiments show that our approach can substantially outperform existing methods and generate comments with more novel and valuable information.
%
%
%
\nocite{ctga}
\bibliographystyle{splncs04}
\bibliography{nlpcc}

\end{document}